\documentclass{article} % For LaTeX2e
\usepackage{iclr2027_conference,times}

\usepackage{amsmath,amsfonts,bm}

\def\eqref#1{equation~\ref{#1}}
\def\1{\bm{1}}

\DeclareMathAlphabet{\mathsfit}{\encodingdefault}{\sfdefault}{m}{sl}
\SetMathAlphabet{\mathsfit}{bold}{\encodingdefault}{\sfdefault}{bx}{n}

\usepackage{hyperref}
\usepackage{url}
\usepackage[pdftex]{graphicx}
\usepackage{pifont}
\usepackage{booktabs}
\usepackage{multirow}
\usepackage{comment}
\usepackage{wrapfig}
\usepackage[table]{xcolor}

\newcommand{\best}[1]{\cellcolor{gray!18}\ensuremath{\mathbf{#1}}}

\title{FinFraudBench: A Heterogeneous Graph Benchmark for Financial Fraud Detection}

\author{
\textbf{Yixuan Chen$^{1,2}$\thanks{Equal contribution.} \quad Hongyu Zhan$^1$\footnotemark[1] \quad Jie Sheng$^3$ \quad Weiyu Han$^3$}\\
\textbf{Shuai Chen$^3$ \quad Tianyi Zhang$^3$ \quad Xiao Tan$^3$\thanks{Corresponding authors.} \quad Jun Xia$^{1,4}$\footnotemark[2]}\\
\normalfont
$^1$HKUST-GZ \quad
$^2$Jilin University \quad
$^3$Ant Group \quad
$^4$HKUST\\
\texttt{\small chenyx2123@mails.jlu.edu.cn, hzhan701@connect.hkust-gz.edu.cn}\\
\texttt{\small \{shengjie.sheng, hanweiyu.hwy, shuai.cs\}@ant-intl.com}\\
\texttt{\small \{zty113091, alex.tx\}@ant-intl.com}\\
\texttt{\small junxia@hkust-gz.edu.cn}
}

\iclrfinalcopy % Uncomment for camera-ready version, but NOT for submission.
\begin{document}

\maketitle

\begin{abstract}
The increasing complexity of digital financial systems has reshaped financial fraud detection from isolated transaction classification into relational risk reasoning over interconnected financial entities. This shift has motivated graph-based fraud detection, where models identify fraudulent nodes by exploiting dependencies among customers, cards, merchants, categories, and locations. However, despite rapid progress in graph-based methods, existing public benchmarks remain misaligned with real-world financial systems in two important aspects. First, they often simplify financial ecosystems into homogeneous or single-node-type multi-relational graphs, failing to preserve the multi-entity and multi-relational nature of financial data. Second, they rarely provide large-scale heterogeneous financial graph datasets with realistic operating conditions such as extreme class imbalance and limited label availability, making it difficult to assess the practical effectiveness of current methods. To address these gaps, we present \textbf{FinFraudBench}, a heterogeneous graph benchmark for financial fraud detection. FinFraudBench contains two heterogeneous graph datasets (\textbf{CreditCard-Fraud} and \textbf{BankTrans-Fraud}) with up to 8.99M nodes and 89.23M directed typed edges. Each dataset preserves six financial entity types, fourteen directed edge types, and natural fraud rates that mirror deployment constraints. With these datasets, we establish a standardized evaluation protocol covering both ranking and imbalance-sensitive classification metrics, and evaluate representative baselines. Extensive experiments yield empirical insights into current methods' limitations and suggest promising avenues for future research. FinFraudBench is available at \url{https://anonymous.4open.science/r/FinFraudBench-B002}. 
\end{abstract}

\section{Introduction}

Financial fraud detection is a central task in modern financial systems, where fraud can cause substantial economic loss and erode trust in digital payment services~\cite{bolton2002statistical,phua2010survey}. Fraud is rarely an isolated property of a single record. A suspicious transaction often becomes easier to identify when considered together with related financial entities and their interactions, such as the customer behind the transaction, the payment instrument, the receiving merchant, and the associated context. This relational nature has motivated graph-based fraud detection, where models can capture dependencies that are difficult to represent from tabular records alone in fraud analysis~\cite{akoglu2015graph,ma2021graphanomaly,liu2022bond}.

\begin{figure*}[!t]
  \centering
  \includegraphics[width=0.88\textwidth]{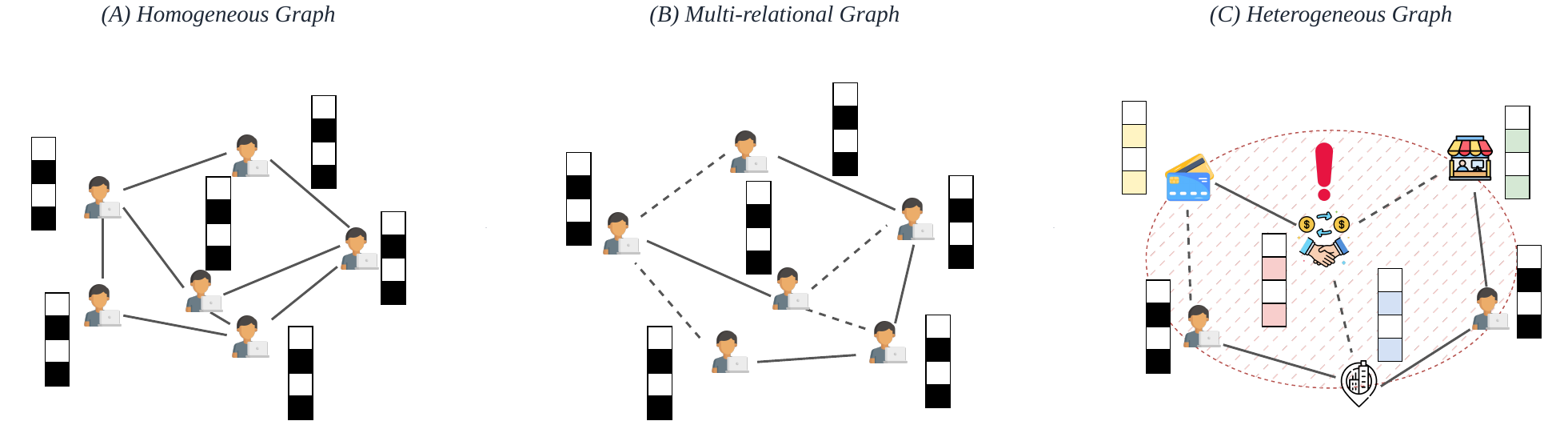}
  \caption{Comparison of three graph formulations for financial fraud detection. Homogeneous graphs use one node type and one relation type; multi-relational graphs keep a single target node type while distinguishing multiple relation types; heterogeneous graphs preserve multiple entity types, multiple relation types, and type-specific attributes.}
  % Description: A three-part schematic comparing homogeneous graphs, multi-relational graphs, and heterogeneous graphs for financial fraud detection.
  \label{fig:graph-formulation-overview}
    \vspace{-8pt}
\end{figure*}

Existing public graph benchmarks for fraud detection, however, often simplify this financial ecosystem before learning begins. We distinguish three graph formulations in this paper. Here, each node represents an entity instance, such as a transaction, user, account, card, merchant, or location. In a homogeneous graph, transactions, users, or accounts are usually selected as the only node type, and all edges describe one broad relation, such as a transaction between users or a payment flow between transactions~\cite{weber2019elliptic,huang2022dgraph}. A multi-relational graph still keeps a single target node type, but its relation types are often induced by typed meta paths in the original heterogeneous ecosystem, such as two target nodes connected through a shared device, shared account, shared card, or shared merchant~\cite{shi2017hinsurvey,dong2017metapath2vec,rayana2015collective,mcauley2013amateurs,liu2020graphconsis,dou2020caregnn}. A heterogeneous graph contains multiple node types and multiple relation types, allowing financial entities and their interactions to be represented directly~\cite{shi2017hinsurvey,hu2020hgt}. Homogeneous and multi-relational formulations are useful abstractions and have supported important progress in graph-based fraud detection, but both lose part of the fraud entity-relation structure.

This representation gap matters for benchmark design. Financial transactions naturally involve multiple entity types, multiple relation semantics, and type-specific attribute spaces. When typed entities are folded into features, converted into target--target edges, or omitted, benchmarks no longer directly test entity- and relation-aware fraud modeling. Figure~\ref{fig:graph-formulation-overview} clearly illustrates this distinction among homogeneous graphs, multi-relational graphs, and heterogeneous graphs considered in this work. To address this gap, we present FinFraudBench, a benchmark consisting of two heterogeneous graph datasets constructed from public financial fraud data sources~\cite{kaggle_kartik_fraud_detection,kaggle_computingvictor_transactions_fraud}; the original sources are transaction-level tables rather than pre-existing graphs. To the best of our knowledge, FinFraudBench is the first benchmark to construct comparatively large heterogeneous graphs specifically for financial fraud detection from public transaction-level sources. Both datasets use a unified schema with transaction nodes as prediction targets and additional typed financial entities as contextual information. FinFraudBench releases the heterogeneous graph as the canonical representation and supports deterministic projections for feature-only, homogeneous, or multi-relational methods.

FinFraudBench also provides a reproducible evaluation protocol. Each dataset is organized into labeled training transactions, unlabeled training transactions, validation transactions, a full test split, and balanced mini-test subsets, supporting controlled comparison under limited labels and extreme class imbalance in realistic deployment settings. We further reproduce representative baselines across non-GNN, homogeneous GNN, multi-relation, fraud-oriented, and heterogeneous families. The results reveal a clear family-level pattern: methods designed for heterogeneous graphs are strongest overall, multi-relation and fraud-oriented methods remain competitive, and target-only non-GNN baselines are limited by their inability to use relational financial context encoded by the benchmark graph structure during practical financial fraud detection tasks.

The main contributions of this work are summarized as follows:
\begin{itemize}
    \item \textbf{Heterogeneous financial fraud graph benchmarks.} We construct two heterogeneous graph datasets from public transaction-level fraud sources. The datasets preserve multiple financial entity types, multiple relation types, and type-specific attributes, while also supporting deterministic simplified views for methods that require homogeneous or multi-relational graph inputs.
    \item \textbf{Reproducible limited-label evaluation protocol.} We
    provide unified graph schemas, labeled and unlabeled training
    transactions, validation splits, balanced mini-test subsets, and
    standardized metrics to support controlled benchmarking under
    label scarcity and extreme class imbalance in a unified evaluation
    setting.
    \item \textbf{Empirical insights for heterogeneous fraud detection.} Our experiments show that heterogeneous graph structures preserve rich signals encoded in typed financial entities and relations, which are essential for fraud modeling. These results point to a promising direction: extending fraud-oriented designs to heterogeneous architectures, so that typed structural information and domain-specific fraud cues can be jointly exploited.
\end{itemize}

\section{Preliminaries and Related Work}

\subsection{Preliminaries}

\textbf{Homogeneous graphs.}
We define the three graph formulations used in this paper. A homogeneous attributed graph is written as $G=(\mathcal{V},\mathcal{E},\mathbf{X})$, where $\mathcal{V}$ is the node set, $\mathcal{E}$ is the edge set, and $\mathbf{X}\in\mathbb{R}^{|\mathcal{V}|\times d}$ is a shared node feature matrix. It assumes one node type and one edge type.

\textbf{Multi-relational graphs.}
A multi-relational graph is a single-node-type graph with multiple relation types, written as $G=(\mathcal{V},\mathcal{E},\mathcal{R},\mathbf{X},\phi)$, where $\mathcal{R}$ is the relation-type set and $\phi:\mathcal{E}\rightarrow\mathcal{R}$ assigns each edge to a relation type. Equivalently, $\mathcal{E}=\bigcup_{r\in\mathcal{R}}\mathcal{E}_{r}$ with $\mathcal{E}_{r}=\{e\in\mathcal{E}:\phi(e)=r\}$. This formulation captures relation heterogeneity while retaining a single node type.

\textbf{Heterogeneous graphs.}
A heterogeneous graph is written as $G=(\mathcal{V},\mathcal{E},\mathcal{A},\mathcal{R},\{\mathbf{X}^{a}\}_{a\in\mathcal{A}},\tau,\phi)$, where $\mathcal{A}$ is the node-type set, $\mathcal{R}$ is the relation-type set, $\tau:\mathcal{V}\rightarrow\mathcal{A}$ assigns each node to a node type, and $\phi:\mathcal{E}\rightarrow\mathcal{R}$ assigns each edge to a relation type. For each node type $a\in\mathcal{A}$, nodes of that type are collected as $\mathcal{V}_{a}=\{v\in\mathcal{V}:\tau(v)=a\}$ and have their own feature matrix $\mathbf{X}^{a}\in\mathbb{R}^{|\mathcal{V}_{a}|\times d_a}$. In our constructed benchmark, this stricter heterogeneous graph setting captures both relation-level and entity-level heterogeneity.

\textbf{Fraud detection task.} Given a graph $G$ in one of the above formulations, we formulate financial fraud detection as binary node classification over a target node type. Let $a^{\star}$ be the target type, $\mathcal{V}_{a^{\star}}$ be its nodes, and $\mathcal{V}_{L}\subseteq\mathcal{V}_{a^{\star}}$ be the labeled training nodes. Unlabeled target nodes may remain in the graph as relational context but do not contribute to the supervised loss. Given labels $y_v\in\{0,1\}$, a model assigns each target node a fraud score
\begin{equation}
s_v=f_{\theta}(G,v), \quad \hat{y}_v=\mathbf{1}[s_v\geq \gamma],
\end{equation}
where $\gamma$ is a decision threshold. The supervised objective is
\begin{equation}
\theta^{\star}=\arg\min_{\theta}\sum_{v\in\mathcal{V}_{L}}
\ell\big(s_v,y_v\big).
\end{equation}
In FinFraudBench, the target type is transaction, while customers, cards, merchants, categories, and locations provide relational context.
\subsection{Related Work}

\textbf{Graph-based fraud detection.} Fraud detection has been studied using tabular transaction features, anomaly detection, and graph-based learning~\cite{bolton2002statistical,phua2010survey,chandola2009anomaly,breunig2000lof,liu2008iforest,akoglu2015graph,ma2021graphanomaly}. General GNN backbones provide message-passing, convolutional, inductive, attention-based, and scalable neighborhood aggregation mechanisms~\cite{scarselli2009gnn,gilmer2017mpnn,kipf2017gcn,hamilton2017graphsage,velickovic2018gat,chiang2019clustergcn,zeng2020graphsaint,wu2019gnnsurvey}. In fraud detection, graph-based methods represent suspicious behaviors as relational structures among transactions, accounts, users, reviews, devices, and payment flows, while fraud-oriented GNNs further introduce assumptions about noisy neighborhoods, camouflage, label scarcity, partitioned propagation, or attribute association~\cite{liu2020gem,liu2020graphconsis,dou2020caregnn,chen2024consisgad,zhuo2024pmp,duan2025gaap,li2025dgp}. These works motivate relational modeling for fraud detection, but their public evaluation settings are usually built on homogeneous graphs or single-node-type multi-relational graphs.

\textbf{Heterogeneous graph learning.} Heterogeneous graph learning broadly studies graphs with typed nodes, typed edges, or both. To avoid ambiguity, we use multi-relational graphs for the single-node-type setting with multiple relation types, and heterogeneous graphs for the setting with multiple node types and multiple relation types. Heterogeneous information network research formalizes entity types, relation types, and meta-path semantics~\cite{shi2017hinsurvey,dong2017metapath2vec}, while neural models learn relation-specific transformations, meta-path-based attention, learned composite relations, and node- and edge-type-dependent attention~\cite{schlichtkrull2018rgcn,wang2019han,fu2020magnn,yun2019gtn,hu2020hgt,yang2023sehgnn}. These methods are relevant to financial fraud detection because financial risk data naturally involve different entity roles and relation semantics, yet public fraud benchmarks rarely expose these elements together with type-specific attributes.

\textbf{Fraud datasets and benchmarks.} Public benchmarks strongly influence how fraud detection models are designed and evaluated. YelpChi and Amazon are widely used in graph-based fraud detection; although not financial datasets, they test whether models handle noisy and suspicious relational patterns~\cite{rayana2015collective,mcauley2013amateurs,zhang2020graphrfi}. In common graph-fraud settings, however, they are usually treated as single-node-type multi-relational graphs, where all nodes share the same prediction target type and edge types encode alternative relations between target nodes~\cite{liu2020graphconsis,dou2020caregnn}. Financial fraud benchmarks such as Elliptic and DGraph are closer to our domain, but they are representative homogeneous graphs: Elliptic connects Bitcoin transaction nodes through payment flows, while DGraph models financial users with temporal interaction edges, strong class imbalance, and many unlabeled nodes~\cite{weber2019elliptic,huang2022dgraph}. Recent efforts such as H$^2$GB construct large-scale heterogeneous graphs across domains, highlighting the value of realistic structural properties, but they are not specifically designed for fraud detection~\cite{lin2024h2gb}. Benchmark studies further show that dataset structure shapes the modeling assumptions methods are expected to exploit~\cite{hu2020ogb,dwivedi2020benchmarking,lv2021hgb,lin2024h2gb}. The lack of public financial fraud benchmarks with heterogeneous graph structure therefore limits the development and evaluation of models that reason over multiple financial entity types, relation types, and type-specific attributes. FinFraudBench fills this gap by providing a unified benchmark with two heterogeneous graph datasets constructed from public transaction-level sources and standardized splits for node-level fraud detection.
\section{Benchmark Construction}
\label{sec:construction}
\subsection{Source Data and Construction Principles}
\label{sec:source-design}
\begin{figure}[!t]
  \centering
  \includegraphics[width=0.95\linewidth]{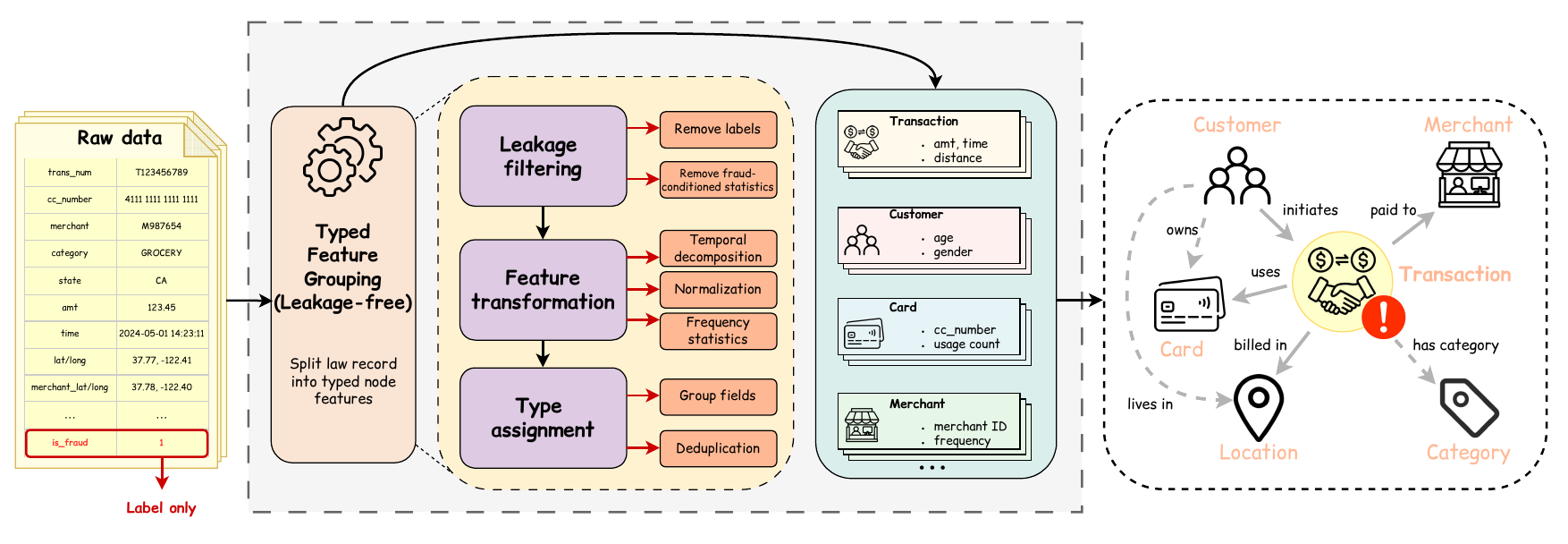}
  \caption{Overview of the graph construction pipeline. Public transaction-level tables are converted into heterogeneous graphs by mapping transactions and recurring financial objects to typed nodes, deriving semantic relations between observed entities, constructing type-specific attributes, and assigning fraud labels only to transaction nodes.}
  % Description: Pipeline for constructing heterogeneous fraud graphs from transaction-level tables.
  \label{fig:graph-construction-pipeline}
  \vspace{-8pt}
\end{figure}

We construct CreditCard-Fraud and BankTrans-Fraud from two public tabular fraud datasets~\cite{kaggle_kartik_fraud_detection,kaggle_computingvictor_transactions_fraud}. Both raw sources are transaction-level tables: each row corresponds to one transaction and contains a fraud label together with non-label fields describing the transaction itself, such as amount and time, and the observable financial context around it, such as customer- or payment-instrument identifiers, merchant and category information, and geographic or billing attributes. These fields are provided as columns in transaction records rather than as separate entity or relation tables. Our goal is to convert such tabular records into heterogeneous graphs while avoiding label leakage in graph or feature construction. Figure~\ref{fig:graph-construction-pipeline} summarizes this construction pipeline for both datasets.

The construction follows three principles. First, recurring financial objects are represented as typed nodes rather than being kept only as transaction attributes. Second, relation types follow the semantic roles between typed entities, so different entity associations are preserved as distinct edge types. Third, labels remain defined only on transaction nodes, while non-transaction nodes provide unlabeled context for limited-label fraud detection.

\subsection{Graph Schema, Attributes, and Labels}
\label{sec:schema-construction}
Both datasets share a transaction-centered heterogeneous schema with six node types and seven semantic edge types. Transaction nodes are the prediction targets, while customer, card, merchant, category, and location nodes provide typed context, where location denotes state or location fields in the source records. The seven edge types include five transaction-context associations and two context-context associations, customer--card ownership and customer--location association. Adding reverse edges yields fourteen directed edge types.

\begin{comment}
\begin{figure}[!t]
  \centering
  \includegraphics[width=0.5\linewidth]{Figures/appendix_figure.pdf}
  \caption{Local visualization of the constructed heterogeneous graph. The typed relations define transaction-centered meta-paths through customers, cards, merchants, categories, and locations. For readability, the illustration shows one direction for each semantic relation; the actual benchmark additionally stores the corresponding reverse edge type for directed message passing.}
  % Description: A local heterogeneous graph visualization showing transaction, customer, card, merchant, category, and location nodes connected by typed relations.
  \label{fig:constructed-graph-local-view}
\end{figure}
\end{comment}
Each source-table row is mapped to one transaction node, with entity-valued fields deduplicated within their own node types. The induced typed edges preserve intermediate entities that link transactions, rather than projecting shared contexts into direct transaction--transaction edges. Node attributes are built from non-label source fields and leakage-free deterministic features, with type-specific dimensions ranging from 7 to 21. Fraud labels are attached only to transaction nodes; all other node types provide unlabeled relational context. A compact schema and feature summary is provided in Appendix~\ref{app:additional-details}.

\section{Dataset Statistics and Analysis}
\label{sec:analysis}

\subsection{Graph Scale and Heterogeneity}
\label{sec:scale-heterogeneity}

Table~\ref{tab:graph-scale-statistics} summarizes the graph-scale and label statistics of the two datasets. CreditCard-Fraud contains 1.86M nodes and 18.53M directed typed edges, while the larger BankTrans-Fraud contains 8.99M nodes and 89.23M directed typed edges. Both datasets share the same schema of six node types and fourteen directed edge types, but differ substantially in scale and fraud prevalence. Although transaction nodes dominate graph size, the non-transaction context nodes are central to the relational structure: they represent customers, cards, merchants, categories, and locations, and link transactions through shared financial entities and contexts instead of collapsing these connections into direct transaction--transaction edges.

\begin{comment}
\begin{table}[!t]
\centering
\caption{Node counts and feature dimensions by node type.}
\label{tab:node-feature-dims}
\footnotesize
\setlength{\tabcolsep}{4.2pt}
\renewcommand{\arraystretch}{1.04}
\begin{tabular}{llrr}
\toprule
Dataset & Node type & \#Nodes & Feat. dim. \\
\midrule
\multirow{6}{*}{\shortstack{CreditCard\\-Fraud}} & transaction & 1.85M & 16 \\
& customer & 999 & 16 \\
& card & 999 & 11 \\
& merchant & 693 & 10 \\
& category & 14 & 7 \\
& state & 51 & 10 \\
\midrule
\multirow{6}{*}{\shortstack{BankTrans\\-Fraud}} & transaction & 8.91M & 17 \\
& customer & 1.22K & 20 \\
& card & 4.07K & 21 \\
& merchant & 66.54K & 10 \\
& category & 109 & 7 \\
& state & 200 & 8 \\
\bottomrule
\end{tabular}
\end{table}
\end{comment}

\subsection{Label Imbalance and Splits}
\label{sec:label-split-analysis}
\begin{wrapfigure}{r}{0.42\linewidth}
  \vspace{-0.8em}
  \centering
  \includegraphics[width=0.98\linewidth]{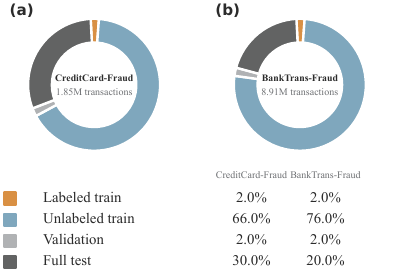}
  \caption{Transaction split proportions of the constructed fraud graphs.}
  \label{fig:dataset-statistics}
  \vspace{-0.8em}
\end{wrapfigure}
Figure~\ref{fig:dataset-statistics} visualizes the main transaction split proportions. Each dataset is organized into labeled training transactions, unlabeled training transactions, validation transactions, a full test split, and five balanced mini-test subsets. The labeled training portion is intentionally small overall, about 2\% of transactions in each dataset: CreditCard-Fraud contains 37.0K labeled and 1.22M unlabeled training transactions, while BankTrans-Fraud contains 178.3K labeled and 6.78M unlabeled training transactions. Unlabeled training transactions remain in the graph as context, so semi-supervised and transductive methods can use their node attributes and local neighborhoods without observing their labels. The full test set preserves the natural class distribution for realistic evaluation, while balanced mini-test subsets contain equal numbers of fraudulent and non-fraudulent transactions for broader and more stable comparison under extreme class imbalance in deployment-like settings.

\begin{table}[!t]
\centering
\caption{Graph-scale and label statistics of the constructed datasets.}
\label{tab:graph-scale-statistics}
\scriptsize
\setlength{\tabcolsep}{2.6pt}
\renewcommand{\arraystretch}{1.02}
\resizebox{\linewidth}{!}{%
\begin{tabular}{lrrrrrr}
\toprule
Dataset & \#Trans. nodes & \#Total nodes & \#Context nodes & \#Directed edges & \#Fraud trans. & Fraud rate \\
\midrule
CreditCard-Fraud & 1.85M & 1.86M & 2.76K & 18.53M & 9,651 & 0.5210\% \\
BankTrans-Fraud & 8.91M & 8.99M & 72.14K & 89.23M & 13,332 & 0.1495\% \\
\bottomrule
\end{tabular}%
}
\end{table}

\section{Benchmark Tasks and Evaluation Protocol}
\label{sec:evaluation-protocol}

The benchmark evaluates transaction-level fraud detection under the splits defined in Section~\ref{sec:label-split-analysis}. The canonical input is the heterogeneous graph constructed in Section~\ref{sec:construction}, and deterministic projections support methods that require feature-only, homogeneous, or multi-relational inputs.

\subsection{Task Setup}
\label{sec:task-setup}
Given the full graph or one of its projections, the task is binary node classification on transaction nodes. Models may use the structure and attributes available in their input view, but supervised labels are provided only for labeled training transactions. Validation labels are used only for model selection and early stopping.

\subsection{Metrics and Reporting Protocol}
\label{sec:metrics-protocol}
Our main experiments use the balanced mini-test subsets defined in Section~\ref{sec:label-split-analysis} to compare ranking and classification behavior under equal fraud/non-fraud weight. We report six complementary metrics for each method.

\textbf{AUROC and AUPRC.}
AUROC measures whether fraudulent transactions receive higher risk scores than legitimate transactions across thresholds~\cite{fawcett2006roc}. AUPRC reflects the precision--recall trade-off for the fraud class and is especially important under class imbalance~\cite{davis2006prroc,saito2015prroc}. Together, they evaluate score ranking before threshold selection in deployment settings.

\textbf{Accuracy and Macro-F1.}
Accuracy is usually misleading on naturally imbalanced fraud data, but is informative on our balanced mini-test subsets because both classes contribute equally. Macro-F1 averages the fraud and non-fraud F1 scores and is more sensitive to failures on either class.

\textbf{Fraud F1 and Fraud Recall.}
Fraud F1 summarizes positive-class detection while penalizing excessive false positives. Fraud Recall measures the fraction of detected fraud cases and captures missed-fraud risk, but should be interpreted with AUPRC and Fraud F1.

\textbf{Reporting protocol.}
In the main benchmark tables, we report mean performance over five balanced mini-test subsets. Unless otherwise stated, model selection uses validation AUPRC. AUROC and AUPRC are ranking metrics, while accuracy, Macro-F1, Fraud F1, and Fraud Recall are threshold-dependent classification metrics.

\section{Experiments}
\label{sec:experiments}

We evaluate representative baselines that cover different modeling assumptions on the same benchmark. We organize methods into five comparison families. MLP and the target-only LLM prompting probes are non-GNN baselines. GCN and GraphSAGE represent homogeneous graph learning on collapsed graph views~\cite{kipf2017gcn,hamilton2017graphsage}. R-GCN represents relation-aware message passing around transaction nodes~\cite{schlichtkrull2018rgcn}; ConsisGAD is grouped with multi-relation methods because its evaluated implementation explicitly uses relation-specific graph channels on multi-relation fraud graphs~\cite{chen2024consisgad}. PMP~\cite{zhuo2024pmp} and GAAP~\cite{duan2025gaap} represent fraud-oriented graph learning methods with assumptions tailored to graph fraud or anomaly detection. HAN, SeHGNN, and HGT represent heterogeneous graph learning methods with typed relation, meta-path, semantic, or node- and edge-type-aware message passing~\cite{wang2019han,yang2023sehgnn,hu2020hgt}. The two LLM probes use Qwen2-7B-Instruct~\cite{yang2024qwen2}: zero-shot prompting~\cite{brown2020language,kojima2022large} and 4-shot in-context prompting~\cite{brown2020language,min2022rethinking}. These probes serve as additional non-GNN references in our benchmark.

\subsection{Overall Benchmark Results}
\label{sec:overall-results}
\begin{figure*}[!t]
  \centering
  \includegraphics[width=0.85\textwidth]{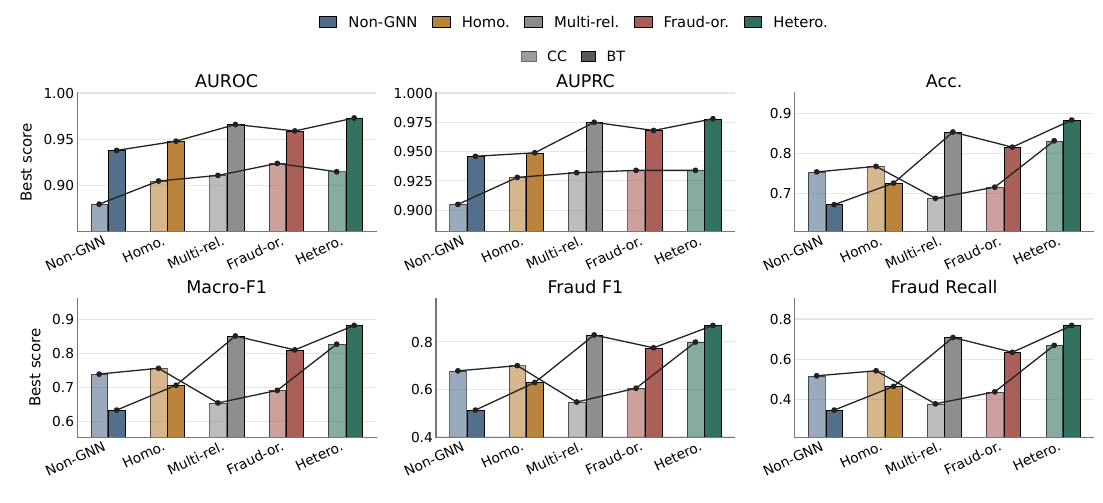}
  \caption{Family-level best results on balanced mini-test subsets. For each metric and model family, bars report the best-performing method within that family in Table~\ref{tab:main-results}; CC denotes CreditCard-Fraud and BT denotes BankTrans-Fraud.}
  % Description: Six-panel grouped bar chart comparing model families on AUROC, AUPRC, accuracy, Macro-F1, Fraud F1, and Fraud Recall for CreditCard-Fraud and BankTrans-Fraud.
  \label{fig:family-best-comparison}
\end{figure*}

Table~\ref{tab:main-results} reports method-level results averaged over five balanced mini-test subsets, while Figure~\ref{fig:family-best-comparison} summarizes the best method in each family for every dataset and metric. We report AUROC, AUPRC, accuracy, Macro-F1, Fraud F1, and Fraud Recall, grouping methods into non-GNN, homogeneous GNN, multi-relation, fraud-oriented, and heterogeneous families.

\begin{table*}[!t]
\centering
\caption{Overall benchmark results on balanced mini-test subsets. Results are averaged over five mini-test subsets and reported as mean $\pm$ standard deviation. The best result for each dataset and metric is highlighted with boldface and gray shading for visual emphasis.}
\label{tab:main-results}
\small
\setlength{\tabcolsep}{2.05pt}
\renewcommand{\arraystretch}{1.04}
\resizebox{0.93\textwidth}{!}{%
\begin{tabular}{lllcccccc}
\toprule
Dataset & Method & Model family & AUROC & AUPRC & Acc. & Macro-F1 & Fraud F1 & Fraud Recall \\
\midrule
\multirow{12}{*}{CreditCard-Fraud}
& MLP & Non-GNN & $0.880{\pm}0.006$ & $0.905{\pm}0.005$ & $0.754{\pm}0.005$ & $0.739{\pm}0.006$ & $0.678{\pm}0.009$ & $0.517{\pm}0.010$ \\
& LLM-ZS & Non-GNN & $0.659{\pm}0.005$ & $0.688{\pm}0.009$ & $0.551{\pm}0.003$ & $0.452{\pm}0.004$ & $0.219{\pm}0.006$ & $0.126{\pm}0.004$ \\
& LLM-ICL-4 & Non-GNN & $0.807{\pm}0.042$ & $0.850{\pm}0.036$ & $0.684{\pm}0.012$ & $0.650{\pm}0.016$ & $0.541{\pm}0.026$ & $0.374{\pm}0.025$ \\
& GCN & Homogeneous GNN & $0.820{\pm}0.011$ & $0.863{\pm}0.007$ & $0.542{\pm}0.003$ & $0.420{\pm}0.007$ & $0.155{\pm}0.012$ & $0.084{\pm}0.007$ \\
& GraphSAGE & Homogeneous GNN & $0.905{\pm}0.006$ & $0.928{\pm}0.004$ & $0.768{\pm}0.010$ & $0.756{\pm}0.012$ & $0.700{\pm}0.017$ & $0.542{\pm}0.020$ \\
& R-GCN & Multi-relation & $0.887{\pm}0.007$ & $0.916{\pm}0.004$ & $0.688{\pm}0.007$ & $0.654{\pm}0.009$ & $0.547{\pm}0.014$ & $0.377{\pm}0.013$ \\
& ConsisGAD & Multi-relation & $0.911{\pm}0.004$ & $0.932{\pm}0.003$ & $0.653{\pm}0.006$ & $0.605{\pm}0.008$ & $0.468{\pm}0.014$ & $0.305{\pm}0.012$ \\
& PMP & Fraud-oriented & $0.868{\pm}0.008$ & $0.903{\pm}0.005$ & $0.699{\pm}0.007$ & $0.670{\pm}0.009$ & $0.571{\pm}0.013$ & $0.400{\pm}0.012$ \\
& GAAP & Fraud-oriented & \best{0.924{\pm}0.004} & \best{0.934{\pm}0.003} & $0.716{\pm}0.008$ & $0.691{\pm}0.010$ & $0.605{\pm}0.014$ & $0.437{\pm}0.014$ \\
& HAN & Heterogeneous & $0.903{\pm}0.004$ & $0.927{\pm}0.002$ & $0.686{\pm}0.008$ & $0.652{\pm}0.011$ & $0.542{\pm}0.017$ & $0.372{\pm}0.016$ \\
& SeHGNN & Heterogeneous & $0.903{\pm}0.005$ & $0.928{\pm}0.003$ & $0.758{\pm}0.006$ & $0.743{\pm}0.007$ & $0.681{\pm}0.011$ & $0.517{\pm}0.014$ \\
& HGT & Heterogeneous & $0.915{\pm}0.004$ & \best{0.934{\pm}0.004} & \best{0.832{\pm}0.008} & \best{0.827{\pm}0.008} & \best{0.799{\pm}0.011} & \best{0.669{\pm}0.014} \\
\midrule
\multirow{12}{*}{BankTrans-Fraud}
& MLP & Non-GNN & $0.938{\pm}0.004$ & $0.946{\pm}0.004$ & $0.672{\pm}0.003$ & $0.633{\pm}0.004$ & $0.513{\pm}0.007$ & $0.346{\pm}0.006$ \\
& LLM-ZS & Non-GNN & $0.809{\pm}0.007$ & $0.813{\pm}0.007$ & $0.517{\pm}0.004$ & $0.372{\pm}0.008$ & $0.070{\pm}0.013$ & $0.036{\pm}0.007$ \\
& LLM-ICL-4 & Non-GNN & $0.873{\pm}0.019$ & $0.860{\pm}0.012$ & $0.541{\pm}0.032$ & $0.418{\pm}0.063$ & $0.151{\pm}0.111$ & $0.086{\pm}0.068$ \\
& GCN & Homogeneous GNN & $0.948{\pm}0.003$ & $0.949{\pm}0.003$ & $0.726{\pm}0.005$ & $0.706{\pm}0.007$ & $0.629{\pm}0.010$ & $0.464{\pm}0.011$ \\
& GraphSAGE & Homogeneous GNN & $0.926{\pm}0.002$ & $0.930{\pm}0.003$ & $0.546{\pm}0.004$ & $0.429{\pm}0.008$ & $0.170{\pm}0.014$ & $0.093{\pm}0.009$ \\
& R-GCN & Multi-relation & $0.962{\pm}0.004$ & $0.972{\pm}0.003$ & $0.854{\pm}0.008$ & $0.851{\pm}0.009$ & $0.829{\pm}0.011$ & $0.708{\pm}0.016$ \\
& ConsisGAD & Multi-relation & $0.966{\pm}0.003$ & $0.975{\pm}0.002$ & $0.822{\pm}0.006$ & $0.816{\pm}0.007$ & $0.784{\pm}0.009$ & $0.644{\pm}0.013$ \\
& PMP & Fraud-oriented & $0.959{\pm}0.004$ & $0.968{\pm}0.003$ & $0.816{\pm}0.009$ & $0.810{\pm}0.010$ & $0.775{\pm}0.013$ & $0.634{\pm}0.018$ \\
& GAAP & Fraud-oriented & $0.949{\pm}0.005$ & $0.954{\pm}0.004$ & $0.643{\pm}0.004$ & $0.591{\pm}0.006$ & $0.445{\pm}0.010$ & $0.286{\pm}0.008$ \\
& HAN & Heterogeneous & $0.969{\pm}0.003$ & $0.977{\pm}0.002$ & $0.881{\pm}0.004$ & $0.880{\pm}0.004$ & $0.865{\pm}0.006$ & $0.763{\pm}0.009$ \\
& SeHGNN & Heterogeneous & $0.969{\pm}0.002$ & \best{0.978{\pm}0.001} & \best{0.884{\pm}0.003} & \best{0.882{\pm}0.003} & \best{0.869{\pm}0.004} & \best{0.768{\pm}0.007} \\
& HGT & Heterogeneous & \best{0.973{\pm}0.002} & \best{0.978{\pm}0.001} & $0.779{\pm}0.006$ & $0.768{\pm}0.007$ & $0.717{\pm}0.010$ & $0.559{\pm}0.012$ \\
\bottomrule
\end{tabular}
}

\end{table*}

\begin{comment}
\begin{figure}[!t]
  \centering
  \includegraphics[width=0.78\linewidth]{Figures/figure3.pdf}
  \caption{Average rank summary for all baselines in Table~\ref{tab:main-results}. Lower ranks are better. Ranks are computed within each dataset and metric before being averaged.}
  % Description: Two-panel horizontal bar chart showing average ranks of all baselines for ranking metrics and classification metrics.
  \label{fig:benchmark-average-rank}
\end{figure}
\end{comment}

\subsection{Representation Analysis}
\label{sec:representation-analysis}

\begin{wrapfigure}{r}{0.45\linewidth}
  \vspace{-10pt}
  \centering
  \includegraphics[width=1.0\linewidth]{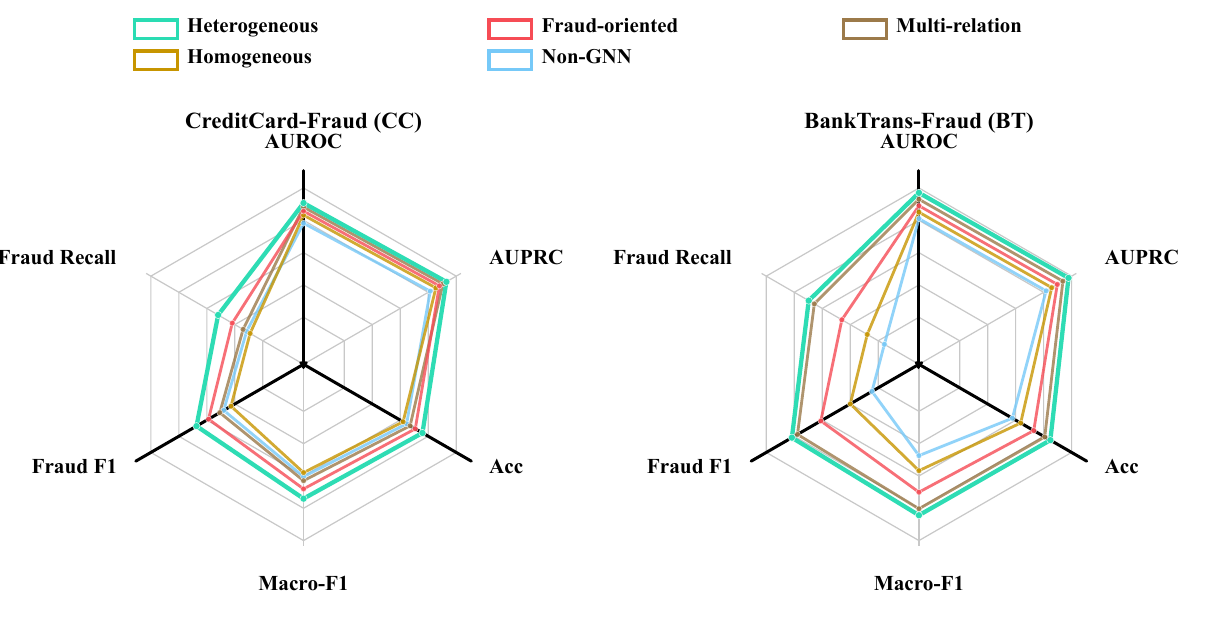}
  \caption{Family-level mean-score radar across benchmark metrics.}
  \label{fig:family-mean-score-radar}
  \vspace{-12pt}
\end{wrapfigure}
Figure~\ref{fig:family-mean-score-radar} complements Figure~\ref{fig:family-best-comparison} with a compact family-level mean-score view across the two datasets. The radar profiles show that heterogeneous methods are strongest overall, while multi-relation and fraud-oriented methods remain competitive and non-GNN baselines are consistently weaker.

\textbf{Transaction features provide a strong but incomplete baseline.} MLP reaches Macro-F1 scores of 0.739 and 0.633 on the two datasets, confirming that target-node attributes carry substantial fraud signals. However, it cannot use the customer, card, merchant, category, and location nodes exposed by the benchmark. As shown in Table~\ref{tab:main-results}, homogeneous propagation is also not consistently beneficial: GCN trails MLP on CreditCard-Fraud, and GraphSAGE improves Macro-F1 there but falls below MLP on the larger graph. This suggests that the main challenge is not merely adding edges, but learning from the heterogeneous relational structure preserved in FinFraudBench.

\textbf{Heterogeneous graph models are strongest overall.} HGT achieves the best accuracy, Macro-F1, Fraud F1, and Fraud Recall on CreditCard-Fraud, while GAAP slightly leads AUROC and AUPRC. On BankTrans-Fraud, HGT obtains the best AUROC and AUPRC, while SeHGNN leads the classification metrics. R-GCN also improves substantially on the larger graph, indicating that relation semantics remain useful even without full node-type heterogeneity. Overall, the results support the benchmark motivation that relation semantics, node types, and type-specific attributes provide complementary fraud signals~\cite{schlichtkrull2018rgcn,wang2019han,yang2023sehgnn,hu2020hgt}.

\begin{figure}[!b]
  \centering
  \includegraphics[width=0.58\linewidth]{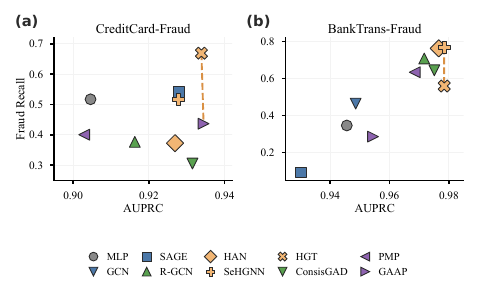}
  \caption{AUPRC--Fraud Recall frontier for trained non-LLM baselines, showing that stronger ranking performance does not always lead to higher fraud recall at the selected decision threshold. Higher values indicate better performance on both axes.}
  % Description: Two-panel scatter plot showing the trade-off between AUPRC and Fraud Recall for trained non-LLM baselines on the two benchmark datasets.
  \label{fig:ranking-recall-frontier}
  \vspace{-8pt}
\end{figure}

\textbf{Relation-aware and fraud-oriented biases remain useful but are not sufficient alone.} GAAP obtains the best AUROC and AUPRC on CreditCard-Fraud, while ConsisGAD and PMP remain competitive on BankTrans-Fraud~\cite{chen2024consisgad,zhuo2024pmp,duan2025gaap}. Their gap to the strongest heterogeneous models on threshold-dependent classification metrics suggests that relation channels or fraud-oriented assumptions should be combined with explicit entity-type modeling. Several methods obtain high AUPRC but differ substantially in Fraud Recall; Figure~\ref{fig:ranking-recall-frontier} illustrates this metric mismatch and supports reporting both ranking metrics and threshold-dependent classification metrics. The LLM prompting baselines are weaker than most graph learning baselines, although 4-shot in-context prompting consistently improves over zero-shot prompting on both datasets, providing a target-only non-GNN reference point for the benchmark.

\subsection{Scalability and Stability Analysis}
\label{sec:scalability-stability-analysis}
Scalability is part of the benchmark rather than a separate engineering detail. CreditCard-Fraud contains 18.53M directed typed edges, and BankTrans-Fraud contains 89.23M, so practical methods must combine predictive quality with sampling or batching strategies. This is especially relevant for graph-based baselines, which can use unlabeled training transactions as context but must do so without full-batch propagation. The balanced mini-test protocol yields stable estimates for most trained baselines across the five subsets, while the two LLM prompting probes show larger standard deviations and are therefore less stable than the graph-learning methods.

\section{Discussion}
\label{sec:discussion}

\subsection{Key Findings}
\label{sec:key-findings}
The experiments suggest that FinFraudBench is useful not only for
ranking baselines, but also for diagnosing which parts of the released
schema a method can exploit. Transaction features, relation channels,
entity types, type-specific attributes, and fraud-oriented assumptions
contribute complementary signals for multi-entity financial fraud
detection.

\textbf{Heterogeneous graph design gives the strongest overall signal.}
Across experiments, heterogeneous methods provide the most consistent gains. This suggests that FinFraudBench exposes useful signals through entity types, relation semantics, and type-specific attributes. Homogeneous or single-node-type multi-relational projections simplify this structure and can discard part of the original fraud entity-relation information.

\textbf{Fraud-oriented designs can build on heterogeneous graphs.}
Multi-relation and fraud-oriented methods remain competitive on
several metrics, showing that relation channels and fraud-oriented
inductive biases are useful for financial fraud detection. However,
their remaining gap to the strongest heterogeneous methods suggests
that these cues are most promising when combined with entity types
and type-specific attributes in heterogeneous financial graphs.

\subsection{Limitations and Future Work}
\label{sec:limitations}
FinFraudBench is constructed from public transaction-level tables, so its graph signals are limited by the fields exposed in those sources and may not fully reflect richer real-world fraud contexts. Future work can extend the benchmark with more complete operational signals and develop fraud-oriented models designed directly for heterogeneous fraud graphs under limited labels and extreme class imbalance in realistic deployment settings.

\section{Conclusion}
\label{sec:conclusion}
We presented FinFraudBench, a benchmark with two heterogeneous graph datasets for financial fraud detection constructed from public transaction-level fraud sources. The proposed graphs preserve multiple financial entity types, relation types, type-specific attributes, limited labels, and realistic fraud imbalance. Our baseline reproduction provides family-level insights into how model designs use heterogeneous fraud signals. The results show that heterogeneous GNNs perform best. Multi-relation and fraud-oriented methods remain competitive but less consistent, highlighting the value of preserving heterogeneous financial structure.

\section*{Data Availability}
\begin{sloppypar}
The original tabular datasets used in this work are publicly available on Kaggle: \url{https://www.kaggle.com/datasets/kartik2112/fraud-detection} and \url{https://www.kaggle.com/datasets/computingvictor/transactions-fraud-datasets}.
\end{sloppypar}

\section*{Ethics and Reproducibility Statement}
\begin{sloppypar}
FinFraudBench is built from public transaction-level data and does not involve human-subject experiments or new data collection. The benchmark is intended for fraud detection research rather than operational decision making without additional validation. All preprocessing, split construction, and reported experiments are described in the paper and appendix. The code repository has been released openly, and the constructed datasets are available on Hugging Face with fixed random splits to support reproduction of the main results.
\end{sloppypar}

\section*{AI Use Statement}
\begin{sloppypar}
We used AI-assisted tools to support literature search and survey, manuscript writing, language polishing, and formatting during paper preparation. These tools were not used to design the proposed benchmark, construct the datasets, define the experimental protocol, or draw scientific conclusions. All scientific claims, dataset construction, experimental analysis, and final wording were carefully reviewed and approved by the authors.
\end{sloppypar}

\clearpage
\bibliography{iclr2027_conference}
\bibliographystyle{iclr2027_conference}

\clearpage
\appendix
\section{Additional Dataset Details}
\label{app:additional-details}
\begingroup
\setlength{\textfloatsep}{6pt plus 1pt minus 2pt}
\setlength{\floatsep}{6pt plus 1pt minus 2pt}
\setlength{\abovecaptionskip}{3pt}
\setlength{\belowcaptionskip}{2pt}

\subsection{Type-Specific Feature Fields}

Table~\ref{tab:app-feature-fields} lists representative non-label feature groups used to build node attributes. The two graphs share the same node and relation schema, but the feature dimension of the same node type can differ across datasets because the two source tables expose different raw fields. In all cases, both graphs avoid label leakage in graph and feature construction.

\begin{table}[!h]
\centering
\caption{Representative non-label feature groups by node type.}
\label{tab:app-feature-fields}
\footnotesize
\setlength{\tabcolsep}{3pt}
\begin{tabular}{p{0.22\columnwidth}p{0.68\columnwidth}}
\toprule
Node type & Representative feature groups \\
\midrule
transaction & amount, time fields, location fields, distance or transaction context \\
customer & profile encodings, location fields, transaction counts and amount statistics \\
card & billing/context encodings, transaction counts, amount statistics, unique entity counts \\
merchant & category encoding, transaction counts, amount statistics, unique customer/card/state counts \\
category & transaction counts, amount statistics, unique merchant/customer/card counts \\
location & transaction counts, amount statistics, unique entity counts, location/population summaries \\
\bottomrule
\end{tabular}
\end{table}

\subsection{Detailed Transaction Splits}

Table~\ref{tab:app-detailed-splits} reports the split statistics behind the compact split summary in the main paper. The unlabeled training split keeps transaction nodes available as graph context while hiding their labels from supervised training.

\begin{table}[!h]
\centering
\caption{Detailed transaction split statistics.}
\label{tab:app-detailed-splits}
\scriptsize
\setlength{\tabcolsep}{4pt}
\begin{tabular}{llrrrr}
\toprule
Dataset & Split & \#Transactions & \#Fraud & \#Benign & Fraud rate \\
\midrule
\multirow{4}{*}{CreditCard-Fraud} & train\_labeled & 37.05K & 214 & 36.83K & 0.5776\% \\
& train\_unlabeled & 1.22M & 7.08K & 1.22M & 0.5789\% \\
& validation & 37.05K & 214 & 36.83K & 0.5776\% \\
& full\_test & 555.72K & 2.15K & 553.57K & 0.3860\% \\
\midrule
\multirow{4}{*}{BankTrans-Fraud} & train\_labeled & 178.30K & 268 & 178.03K & 0.1503\% \\
& train\_unlabeled & 6.78M & 10.13K & 6.77M & 0.1495\% \\
& validation & 178.30K & 268 & 178.03K & 0.1503\% \\
& full\_test & 1.78M & 2.67K & 1.78M & 0.1495\% \\
\bottomrule
\end{tabular}
\end{table}

\clearpage
\subsection{Complete Balanced Mini-Test Results}

Tables~\ref{tab:app-complete-results-credit} and~\ref{tab:app-complete-results-bank} report the complete balanced mini-test results, including Fraud Precision, which is omitted from the main table to keep the main comparison compact.

\begin{table}[!h]
\centering
\vspace{-4pt}
\caption{Complete balanced mini-test results on CreditCard-Fraud. Values are means and standard deviations across five mini-test subsets.}
\label{tab:app-complete-results-credit}
\scriptsize
\setlength{\tabcolsep}{1.75pt}
\resizebox{0.92\textwidth}{!}{%
\begin{tabular}{lrrrrrrr}
\toprule
Method & AUROC & AUPRC & Acc. & Macro-F1 & Fraud F1 & Fraud Prec. & Fraud Recall \\
\midrule
MLP & $0.8798\pm 0.0057$ & $0.9046\pm 0.0052$ & $0.7539\pm 0.0050$ & $0.7393\pm 0.0059$ & $0.6775\pm 0.0087$ & $0.9822\pm 0.0020$ & $0.5172\pm 0.0103$ \\
GCN & $0.8199\pm 0.0107$ & $0.8632\pm 0.0067$ & $0.5416\pm 0.0033$ & $0.4204\pm 0.0065$ & $0.1554\pm 0.0115$ & $0.9861\pm 0.0099$ & $0.0844\pm 0.0068$ \\
GraphSAGE & $0.9046\pm 0.0059$ & $0.9279\pm 0.0044$ & $0.7681\pm 0.0102$ & $0.7556\pm 0.0118$ & $0.7003\pm 0.0167$ & $0.9890\pm 0.0044$ & $0.5422\pm 0.0195$ \\
R-GCN & $0.8870\pm 0.0072$ & $0.9163\pm 0.0044$ & $0.6876\pm 0.0066$ & $0.6542\pm 0.0088$ & $0.5466\pm 0.0137$ & $0.9957\pm 0.0030$ & $0.3768\pm 0.0128$ \\
HAN & $0.9025\pm 0.0038$ & $0.9269\pm 0.0024$ & $0.6858\pm 0.0080$ & $0.6515\pm 0.0108$ & $0.5422\pm 0.0170$ & $0.9978\pm 0.0023$ & $0.3724\pm 0.0160$ \\
SeHGNN & $0.9026\pm 0.0048$ & $0.9279\pm 0.0026$ & $0.7575\pm 0.0059$ & $0.7426\pm 0.0073$ & $0.6807\pm 0.0111$ & $0.9959\pm 0.0041$ & $0.5172\pm 0.0137$ \\
HGT & $0.9150\pm 0.0035$ & $0.9339\pm 0.0036$ & $0.8317\pm 0.0077$ & $0.8271\pm 0.0082$ & $0.7990\pm 0.0106$ & $0.9911\pm 0.0042$ & $0.6694\pm 0.0140$ \\
ConsisGAD & $0.9110\pm 0.0042$ & $0.9315\pm 0.0031$ & $0.6526\pm 0.0058$ & $0.6049\pm 0.0084$ & $0.4676\pm 0.0135$ & $1.0000\pm 0.0000$ & $0.3052\pm 0.0115$ \\
PMP & $0.8683\pm 0.0075$ & $0.9029\pm 0.0053$ & $0.6992\pm 0.0068$ & $0.6697\pm 0.0086$ & $0.5709\pm 0.0129$ & $0.9949\pm 0.0048$ & $0.4004\pm 0.0120$ \\
GAAP & $0.9237\pm 0.0037$ & $0.9344\pm 0.0033$ & $0.7155\pm 0.0077$ & $0.6915\pm 0.0096$ & $0.6054\pm 0.0142$ & $0.9873\pm 0.0062$ & $0.4366\pm 0.0142$ \\
LLM-ZS & $0.6591\pm 0.0046$ & $0.6877\pm 0.0086$ & $0.5510\pm 0.0032$ & $0.4519\pm 0.0039$ & $0.2189\pm 0.0059$ & $0.8412\pm 0.0249$ & $0.1258\pm 0.0036$ \\
LLM-ICL-4 & $0.8067\pm 0.0417$ & $0.8495\pm 0.0360$ & $0.6839\pm 0.0120$ & $0.6501\pm 0.0164$ & $0.5414\pm 0.0260$ & $0.9841\pm 0.0051$ & $0.3739\pm 0.0247$ \\
\bottomrule
\end{tabular}
}
\vspace{-8pt}
\end{table}

\begin{table}[!h]
\centering
\vspace{-4pt}
\caption{Complete balanced mini-test results on BankTrans-Fraud. Values are means and standard deviations across five mini-test subsets.}
\label{tab:app-complete-results-bank}
\scriptsize
\setlength{\tabcolsep}{1.75pt}
\resizebox{0.92\textwidth}{!}{%
\begin{tabular}{lrrrrrrr}
\toprule
Method & AUROC & AUPRC & Acc. & Macro-F1 & Fraud F1 & Fraud Prec. & Fraud Recall \\
\midrule
MLP & $0.9384\pm 0.0043$ & $0.9456\pm 0.0041$ & $0.6718\pm 0.0030$ & $0.6329\pm 0.0042$ & $0.5133\pm 0.0068$ & $0.9926\pm 0.0032$ & $0.3462\pm 0.0063$ \\
GCN & $0.9476\pm 0.0032$ & $0.9486\pm 0.0030$ & $0.7257\pm 0.0052$ & $0.7055\pm 0.0065$ & $0.6285\pm 0.0099$ & $0.9732\pm 0.0051$ & $0.4642\pm 0.0111$ \\
GraphSAGE & $0.9263\pm 0.0019$ & $0.9302\pm 0.0026$ & $0.5460\pm 0.0043$ & $0.4286\pm 0.0082$ & $0.1696\pm 0.0144$ & $0.9915\pm 0.0093$ & $0.0928\pm 0.0086$ \\
R-GCN & $0.9615\pm 0.0039$ & $0.9717\pm 0.0028$ & $0.8542\pm 0.0082$ & $0.8510\pm 0.0087$ & $0.8292\pm 0.0112$ & $1.0000\pm 0.0000$ & $0.7084\pm 0.0164$ \\
HAN & $0.9687\pm 0.0028$ & $0.9766\pm 0.0021$ & $0.8813\pm 0.0043$ & $0.8796\pm 0.0044$ & $0.8654\pm 0.0055$ & $0.9995\pm 0.0012$ & $0.7630\pm 0.0088$ \\
SeHGNN & $0.9689\pm 0.0019$ & $0.9784\pm 0.0012$ & $0.8838\pm 0.0032$ & $0.8822\pm 0.0034$ & $0.8686\pm 0.0041$ & $0.9995\pm 0.0007$ & $0.7680\pm 0.0065$ \\
HGT & $0.9729\pm 0.0020$ & $0.9784\pm 0.0009$ & $0.7793\pm 0.0062$ & $0.7680\pm 0.0071$ & $0.7170\pm 0.0100$ & $0.9989\pm 0.0016$ & $0.5592\pm 0.0116$ \\
ConsisGAD & $0.9657\pm 0.0031$ & $0.9751\pm 0.0019$ & $0.8221\pm 0.0064$ & $0.8163\pm 0.0071$ & $0.7835\pm 0.0095$ & $1.0000\pm 0.0000$ & $0.6442\pm 0.0129$ \\
PMP & $0.9592\pm 0.0036$ & $0.9685\pm 0.0027$ & $0.8163\pm 0.0090$ & $0.8099\pm 0.0099$ & $0.7752\pm 0.0134$ & $0.9978\pm 0.0009$ & $0.6340\pm 0.0179$ \\
GAAP & $0.9493\pm 0.0049$ & $0.9544\pm 0.0042$ & $0.6427\pm 0.0039$ & $0.5907\pm 0.0060$ & $0.4447\pm 0.0099$ & $0.9973\pm 0.0037$ & $0.2862\pm 0.0083$ \\
LLM-ZS & $0.8087\pm 0.0072$ & $0.8127\pm 0.0074$ & $0.5174\pm 0.0041$ & $0.3720\pm 0.0076$ & $0.0697\pm 0.0131$ & $0.9587\pm 0.0504$ & $0.0362\pm 0.0070$ \\
LLM-ICL-4 & $0.8725\pm 0.0186$ & $0.8598\pm 0.0120$ & $0.5406\pm 0.0321$ & $0.4179\pm 0.0628$ & $0.1513\pm 0.1111$ & $0.9561\pm 0.0144$ & $0.0856\pm 0.0678$ \\
\bottomrule
\end{tabular}
}
\vspace{-8pt}
\end{table}

\subsection{Mini-Test Stability}

Figure~\ref{fig:app-mini-test-stability} reports the coefficient of variation across the five balanced mini-test subsets for each metric and trained non-LLM baseline. The figure complements the averaged results in the main paper by showing that the balanced mini-test protocol gives stable estimates for most graph-learning baselines.

\begin{figure}[!h]
  \centering
  \vspace{-4pt}
  \includegraphics[width=0.62\textwidth]{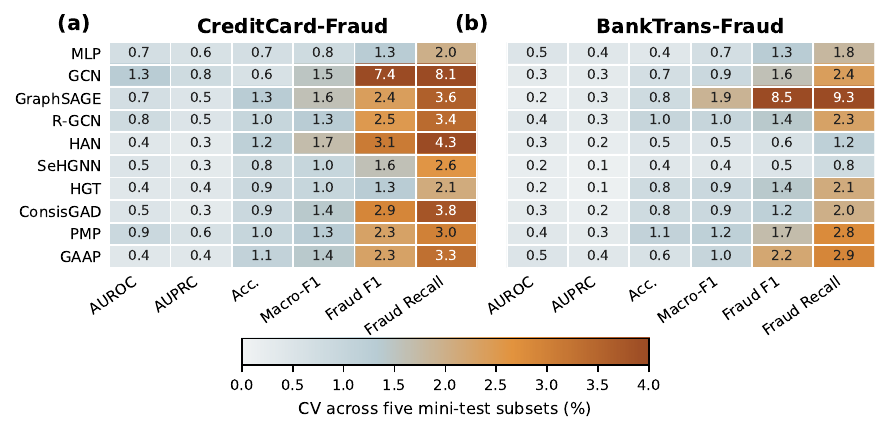}
  \caption{Stability of balanced mini-test evaluation. Heatmaps report the coefficient of variation across five mini-test subsets; lighter cells indicate more stable estimates.}
  \label{fig:app-mini-test-stability}
\vspace{-8pt}
\end{figure}

\endgroup

\end{document}